\documentclass{article}

\usepackage[utf8]{inputenc} % allow utf-8 input
\usepackage[T1]{fontenc}    % use 8-bit T1 fonts
\usepackage{url}            % simple URL typesetting
\usepackage{booktabs}       % professional-quality tables
\usepackage{amsfonts}       % blackboard math symbols
\usepackage{nicefrac}       % compact symbols for 1/2, etc.
\usepackage{microtype}      % microtypography
\usepackage{xcolor}
\usepackage{graphicx}
\usepackage{multirow}
\usepackage{amssymb}
\usepackage{amsmath}
\usepackage{dsfont}
\usepackage{soul}
\usepackage{array}
\usepackage{makecell}

\usepackage{subcaption}
\usepackage{wrapfig}
\DeclareMathOperator*{\argmax}{argmax}

\usepackage{titlesec}

\titlespacing{\paragraph}{%
  0pt}{%              left margin
  0\baselineskip}{% space before (vertical)
  .5em}%               space after (horizontal)

\usepackage{hyperref}

\usepackage[accepted]{icml2020}

\icmltitlerunning{A Generalized Framework of Sequence Generation}

\begin{document}

\twocolumn[
\icmltitle{A Generalized Framework of Sequence Generation with Application to Undirected Sequence Models}



\icmlsetsymbol{equal}{*}

\begin{icmlauthorlist}
\icmlauthor{Elman Mansimov}{nyu}
\icmlauthor{Alex Wang}{nyu}
\icmlauthor{Sean Welleck}{nyu}
\icmlauthor{Kyunghyun Cho}{nyu,fb,cifar}
\end{icmlauthorlist}

\icmlaffiliation{nyu}{New York University}
\icmlaffiliation{fb}{Facebook AI Research}
\icmlaffiliation{cifar}{CIFAR Azrieli Global Scholar}

\icmlcorrespondingauthor{Elman Mansimov}{mansimov@cs.nyu.edu}

\icmlkeywords{Machine Learning, ICML}

\vskip 0.3in
]



\printAffiliationsAndNotice{}  % leave blank if no need to mention equal contribution

\begin{abstract}

Undirected neural sequence models such as BERT \citep{devlin2019bert} have received renewed interest due to their success on discriminative natural language understanding tasks such as question-answering and natural language inference. 
The problem of generating sequences directly from these models has received relatively little attention, in part because generating from undirected models departs significantly from conventional monotonic generation in directed sequence models.
%While undirected neural sequence models such as BERT \citep{devlin2019bert} have received renewed interest due to their successes on a variety of natural language understanding tasks,
%generating sequences directly from these models has received relatively little attention.
We investigate this problem by proposing a generalized model of sequence generation that unifies decoding in directed and undirected models. 
The proposed framework models the process of generation rather than the resulting sequence, and under this framework, we derive various neural sequence models as special cases, such as autoregressive, semi-autoregressive, and refinement-based non-autoregressive models.
This unification enables us to adapt decoding algorithms originally developed for directed sequence models to undirected sequence models.
We demonstrate this by evaluating various handcrafted and learned decoding strategies on a BERT-like machine translation model \citep{lample2019cross}. 
The proposed approach achieves constant-time translation results on par with linear-time translation results from the same undirected sequence model, while both are competitive with the state-of-the-art on WMT'14 English-German translation. 
%Finally, our experiments show that both linear-time and constant-time generation from the same undirected sequence model, under our framework, is competitive with the state of the art on WMT'14 English-German translation. 
%We furthermore observe that the proposed approach enables constant-time translation while remaining within 1 BLEU score compared to linear-time translation from the same undirected neural sequence model.
% , our model 
% %without distillation 
% is competitive with traditional greedy monotonic decoding from a directed model, while performing 92-95 $\%$ relative to beam-search monotonic decoding from a directed model.

%it is competitive with a strong directed baseline, with our algorithm performing best among the considered decoding algorithms for undirected models.
%Qualitatively, we show that the generation order picked by our algorithm flexibly alternates between left-to-right and right-to-left, often switching between these approaches within a single generation.
%\cho{END}
\end{abstract}

\section{Introduction}

Undirected neural sequence models such as BERT \citep{devlin2019bert} have recently brought significant improvements to a variety of discriminative language modeling tasks such as question-answering and natural language inference. Generating sequences from these models has received relatively little attention. Unlike directed sequence models, each word typically depends on the full left and right context around it in undirected sequence models. Thus, a decoding algorithm for an undirected sequence model must specify both how to select positions and what symbols to place in the selected positions. 
We formalize this process of selecting positions and replacing symbols as a general framework of sequence generation, 
%In this paper we formalize this process of decoding by selecting positions and replacing symbols as a generalized framework of sequence generation, 
and unify decoding from both directed and undirected sequence models under this framework. This framing enables us to study generation on its own, independent from the specific parameterization of the sequence models. 

Our proposed framework casts sequence generation as a process of determining the length of the sequence, and then repeatedly alternating between selecting sequence positions followed by generation of symbols for those positions. A variety of sequence models can be derived under this framework by appropriately designing the length distribution, position selection distribution, and symbol replacement distribution. Specifically, we derive popular decoding algorithms such as monotonic autoregressive, non-autoregressive by iterative refinement, and monotonic semi-autoregressive decoding as special cases of the proposed model.
%, monotonic semi-autoregressive, and non-monotonic decoding as special cases of the proposed framework.

This separation of coordinate selection and symbol replacement allows us to build a diverse set of decoding algorithms agnostic to the parameterization or training procedure of the underlying model.
%given a single symbol replacement model, 
We thus fix the symbol replacement distribution as a variant of BERT and focus on deriving novel generation procedures for undirected neural sequence models under the proposed framework. 
%We use a variant of BERT as the symbol replacement distribution, and 
We design a coordinate selection distribution using a log-linear model and a learned model with a reinforcement learning objective to demonstrate that our model generalizes various fixed-order generation strategies, while also being capable of adapting generation order based on the content of intermediate sequences. 

We empirically validate our proposal on machine translation using a translation-variant of BERT called a masked translation model~\citep{lample2019cross}. We design several generation strategies based on features of intermediate sequence distributions
% -- most-to-least likely, least-to-most likely, easy-first and hard-first -- 
and compare them against the state-of-the-art monotonic autoregressive sequence model~\citep{vaswani2017attention} on WMT'14 English-German. Our experiments show that generation from undirected sequence models, under our framework, is competitive with the state of the art, and that adaptive-order generation strategies generate sequences in different ways, including left-to-right, right-to-left and mixtures of these. %Additionally, the reinforcement learning based method learns coordinate selection strategies that match or exceed performance of  hand-crafted strategies.
%\red{Add a sentence about performance of RL based position selection mechanisms} \st{This suggests the potential for designing and learning a more sophisticated coordinate selection mechanism.}

Due to the flexibility in specifying a coordinate selection mechanism, we design constant-time variants of the proposed generation strategies, closely following the experimental setup of \citet{ghazvininejad2019constant}. Our experiments reveal that we can do constant-time translation with the budget as low as 20 iterations (equivalently, generating a sentence of length 20 in the conventional approach) while achieving similar performance to the state-of-the-art-monotonic autoregressive sequence model and linear-time translation from the same masked translation model. This again confirms the potential of the proposed framework and generation strategies. We release the implementation, preprocessed datasets as well as trained models online at \url{https://github.com/nyu-dl/dl4mt-seqgen}.

%\paragraph{Related work: pretraining a generator}
%
%A number of papers have explored using pretrained, undirected language models as parameter initialization scheme for sequence generation.
%\citet{Ramachandran_2017,song2019mass,lample2019cross} use a pretrained undirected language model to initialize a conventional monotonic autoregressive sequence model, while  
%\citet{edunov2019pretrained} use a BERT-like model to initialize the lower layers of a generator however without finetuning it.
%Our work differs from these in that we attempt to directly generate from the pretrained language model, rather than using it as a starting point for learning another sequence model.
% The advantage of our approach is that we can avoid expensive finetuning of these large models, with minimal performance loss.

\section{A Generalized Framework of Sequence Generation}
\label{sec:generalized_sequence_generation}

We propose a generalized framework of probabilistic sequence generation to unify generation from directed and undirected neural sequence models. In this generalized framework, we have a {\it generation sequence} $G$ of pairs of an {\it intermediate sequence} $Y^t=(y^t_1, \ldots, y^t_L)$ and the corresponding {\it coordinate sequence} $Z^t=(z^t_1, \ldots, z^t_L)$, where $V$ is a vocabulary, $L$ is a length of a sequence, $T$ is a number of generation steps, $y^t_i \in V$, and $z^t_i \in \left\{0, 1\right\}$. The coordinate sequence indicates which of the current intermediate sequence are to be replaced. That is, consecutive pairs are related to each other by
% \begin{align*}
    $y^{t+1}_i = 
    (1-z^{t+1}_i) y^t_i + 
    z^{t+1}_i \tilde{y}^{t+1}_i$,
% \end{align*}
where $\tilde{y}^{t+1}_i \in V$ is a new symbol for the position $i$. This sequence of pairs $G$ describes a procedure that starts from an empty sequence \mbox{$Y^1=(\left<\text{mask}\right>, \ldots, \left<\text{mask}\right>)$} and empty coordinate sequence $Z^1=(0, ..., 0)$, iteratively fills in tokens, and terminates after $T$ steps with final sequence $Y^{T}$. 
%This procedure of sequence generation $p(G|X)$ is probabilistically modelled as
We model this procedure probabilistically as $p(G|X)$:
\resizebox{.95\linewidth}{!}{
\begin{minipage}{\linewidth}
\begin{align}
\label{eq:generalized_sequence_generation}
&\underbrace{p(L|X)}_{\text{(c) length predict}}
\prod_{t=1}^T \prod_{i=1}^L
\underbrace{p(z^{t+1}_i | Y^{\leq t}, Z^{t}, X)}_{\text{(a) coordinate selection}}
{\underbrace{p(y^{t+1}_{i}| Y^{\leq t}, X)}_{\text{(b) symbol replacement}}}^{z^{t+1}_i}
\end{align}
\end{minipage}
}
We condition the whole process on an input variable $X$ to indicate that the proposed model is applicable to both conditional and unconditional sequence generation. In the latter case, $X=\emptyset$.

We first predict the length $L$ of a target sequence $Y$ according to $p(L | X)$ distribution to which we refer as (c) length prediction. At each generation step $t$, we first select the next coordinates $Z^{t+1}$ for which the corresponding symbols will be replaced according to $p(z^{t+1}_i | Y^{\leq t}, Z^{t}, X)$, to which we refer as (a) coordinate selection. Once the coordinate sequence is determined, we replace the corresponding symbols according to the distribution $p(y^{t+1}_i|Y^{\leq t}, Z^{t+1}, X)$, leading to the next intermediate sequence $Y^{t+1}$. From this sequence generation framework, we recover the sequence distribution $p(Y|X)$ by marginalizing out all the intermediate and coordinate sequences except for the final sequence $Y^T$. In the remainder of this section, we describe several special cases of the proposed framework, which are monotonic autoregressive, non-autoregressive, semi-autoregressive neural sequence models.% and non-monotonic neural sequence models.

\subsection{Special Cases}

\paragraph{Monotonic autoregressive neural sequence models}

We first consider one extreme case of the generalized sequence generation model, where we replace one symbol at a time, monotonically moving from the left-most position to the right-most. 
%This is a widely-used paradigm of monotonic autoregressive neural sequence model.
In this case, we define the coordinate selection distribution of the generalized sequence generation model in Eq.~\eqref{eq:generalized_sequence_generation}~(a) as
%\begin{align}
%    \label{eq:coordinate_selection_autoregressive}
\mbox{$
    p(z_{i+1}^{t+1} =1 | Y^{\leq t}, Z^{t}, X)
    = \mathds{1}(z^{t}_{i} = 1)
%    % \begin{cases}
%    % 1,& \text{ if } z^{t}_{i} = 1 \\
%    % 0,& \text{ otherwise},
%    % \end{cases}
$},
%\end{align}
where $\mathds{1}(\cdot)$ is an indicator function and $z^1_{1}=1$. This coordinate selection distribution is equivalent to saying that we replace one symbol at a time, shifting from the left-most symbol to the right-most symbol, regardless of the content of intermediate sequences.
We then choose the symbol replacement distribution in Eq.~\eqref{eq:generalized_sequence_generation}~(b) to be
%\begin{align}
%    \label{eq:symbol_replacement_autoregressive}
\mbox{$
    p(y_{i+1}^{t+1} | Y^{\leq t}, X) = p(y_{i+1}^{t+1} | y_{1}^t, y_{2}^t, \ldots, y_{i}^{t}, X),
$}
%\end{align}
for $z_{i+1}^{t+1}=1$. Intuitively, we limit the dependency of $y_{i+1}^{t+1}$ only to the symbols to its left in the previous intermediate sequence $y_{<(i+1)}^t$ and the input variable $X$.
The length distribution \eqref{eq:generalized_sequence_generation}~(c) is implicitly defined by considering how often the special token $\left<\text{eos}\right>$, which indicates the end of a sequence, appears after $L$ generation steps:
% \begin{align*}
\mbox{$
    p(L|X) \propto
    \sum_{y_{1:L-1}}
    \prod_{l=1}^{L-1} p(y_{l+1}^{l+1}=\left< \text{eos} \right>|y_{\leq l}^{\leq l}, X)
$}.
% \end{align*}
% This implies that we do not need the explicit representation of this distribution and can rely on the generation procedure itself to determine the length of each generated sequence.
With these choices, the proposed generalized model reduces to
% \begin{align}
    $p(G|X) = \prod_{i=1}^L p(y_i| y_{<i}, X)$ 
% \end{align}
which is a widely-used monotonic autoregressive neural sequence model. 

\paragraph{Non-autoregressive neural sequence modeling by iterative refinement}
\label{sec:nonauto}

We next consider the other extreme in which we replace the symbols in all positions at every single generation step~\citep{lee2018deterministic}. We design the coordinate selection distribution to be implying that we replace the symbols in all the positions. We then choose the symbol replacement distribution to be as it was in Eq.~\eqref{eq:generalized_sequence_generation}~(b). That is, the distribution over the symbols in the position $i$ in a new intermediate sequence $y_i^{t+1}$ is conditioned on the entire current sequence $Y^t$ and the input variable $X$. We do not need to assume any relationship between the number of generation steps $T$ and the length of a sequence $L$ in this case. The length prediction distribution $p(L|X)$ is estimated from training data.

\paragraph{Semi-autoregressive neural sequence models}

\cite{wang2018semi} recently proposed a compromise between autoregressive and non-autoregresive sequence models by predicting a chunk of symbols in parallel at a time. This approach can also be put under the proposed generalized model. We first extend the coordinate selection distribution of the autoregressive sequence model 
%in Eq.~\eqref{eq:coordinate_selection_autoregressive} 
into
\begin{align*}
    % \label{eq:coordinate_selection_semi}
    &p(z_{k(i+1)+j}^{t+1} = 1 | Y^{\leq t}, Z^{t}, X) =\\
    &= 
    \begin{cases}
    1,& \text{ if } z^{t}_{ki+j} = 1,     \forall j \in \left\{0, 1, \ldots, k \right\} \\
    0,& \text{ otherwise},
    \end{cases}
\end{align*}
where $k$ is a {\it group size}. Similarly we modify the symbol replacement distribution: 
%from Eq.~\eqref{eq:symbol_replacement_autoregressive} to
\begin{align*}
    % \label{eq:symbol_replacement_semi}
    % \mbox{$
    p(y_{k(i+1)+j}^{t+1} | Y^{\leq t}, X) = &p(y_{k(i+1)+j}^{t+1} | y^t_{< k(i+1)}, X), \\
    &\forall j \in \left\{0, 1, \ldots, k \right\},
    % $},
\end{align*}
for $z_i^t=1$. This naturally implies that $T=\left\lceil L/k \right\rceil$. 

% Instead of defining $p(Z^{t+1} | Y^{\leq t}, Z^{t}, X)$ heuristically as it is done in case of monotonic left-to-right, non-autoregressive, and semi-autoregressive neural sequence models, we can approximate this coordinate selection distribution with approximate distribution $q_{\theta}(Z^{t+1} | Y^{\leq t}, Z^{t}, X)$ with learned parameters $\theta$. Since there are factorial number of possible orderings, it is not immediately clear which position targets should be used to learn parameters of distribution $q_{\theta}(Z^{t+1} | Y^{\leq t}, Z^{t}, X)$. \citet{welleck2019nonmonotonic} and \citet{stern2019inserttransformer} assign equal probability to each valid position in the sequence as targets. 
% \citet{welleck2019nonmonotonic} further add coaching method, where a model starts slowly following its own positional preferences during training. 
% \citet{stern2019inserttransformer} bias the model to follow balanced binary tree ordering in order to speed up decoding which ends up being logarithmic in complexity. 
% \citet{gu2019insertiondecoding} bias the model to follow the ordering obtained from beam-search.

%\section{Masked language models as undirected neural sequence models}
\section{Decoding from Masked Language Models}

%Recent work \citep{wang2019bert,ghazvininejad2019constant} has demonstrated that undirected neural sequence models such as BERT~\citep{devlin2019bert} can learn complicated sequence distributions and generate well-formed sequences. In such models, it is relatively straightforward to collect unbiased samples using, for instance, Gibbs sampling. 
%But due to high variance of Gibbs sampling, the generated sequence is not guaranteed to be high-quality relative to a ground-truth sequence. 
%Finding a good sequence in a deterministic manner is also nontrivial.
%Therefore, it is not trivial to find such a good sequence in a deterministic manner. 
In this section, we give an overview of masked language models like BERT,
cast Gibbs sampling under the proposed framework,
and then use this connection to design a set of approximate, deterministic decoding algorithms for undirected sequence models.

\subsection{BERT as an undirected sequence model}

BERT \citep{devlin2019bert} is a masked language model: It is trained to predict a word given the word's left and right context.
Because the model gets the full context, there are no directed dependencies among words, so the model is undirected.
The word to be predicted is masked with a special $\left< \text{mask} \right>$ symbol and the model is trained to predict  
%BERT \citep{devlin2019bert} is a masked language model that is trained to predict a word given both its left and right context. 
%The word to be predicted is masked with a special $\left< \text{mask} \right>$ symbol, which corresponds to 
% the following distribution:
% \begin{align}
% \label{eq:BERT_conditional}
$p(y_i | y_{<i}, \left< \text{mask} \right>, y_{>i}, X)$. We refer to this as the {\it conditional BERT distribution}.
% \end{align}
This objective was interpreted as a stochastic approximation to the pseudo log-likelihood objective \citep{besag1977pll} by \citet{wang2019bert}. 
This approach of full-context generation with pseudo log-likelihood maximization for recurrent networks was introduced earlier by \citet{berglund2015bidirectional}.
%Earlier, \citet{berglund2015bidirectional} introduced the same approach of full-context generation with pseudo log-likelihood maximization for recurrent networks. 
More recently, \citet{sun2017bibs} use it for image caption generation.

% In earlier work, \citet{sun2017bibs} study bidirectional neural sequence generation on fill-in-the-blank image captioning task, where they iterate between running beam-search with separate forward and backward RNNs on the full sequence.

% \cite{wang2019bert} present BERT as a Markov random field language model~\citep{jernite2015mrflm}, where the probability distribution $p(Y|X) = p(y_1, y_2, ..., y_L|X)$ is represented as a complete undirected graph. The objective of predicting the masked out words can be interpreted as a stochastic approximation to the pseudo log-likelihood objective \citep{besag1977pll}.

Recent work \citep{wang2019bert,ghazvininejad2019constant} has demonstrated that undirected neural sequence models like BERT can learn complex sequence distributions and generate well-formed sequences. In such models, it is relatively straightforward to collect unbiased samples using, for instance, Gibbs sampling. 
But due to high variance of Gibbs sampling, the generated sequence is not guaranteed to be high-quality relative to a ground-truth sequence. 
Finding a good sequence in a deterministic manner is also nontrivial.

A number of papers have explored using pretrained language models like BERT to initialize sequence generation models.
\citet{Ramachandran_2017}, \citet{song2019mass}, and \citet{lample2019cross} use a pretrained undirected language model to initialize a conventional monotonic autoregressive sequence model, while  
\citet{edunov2019pretrained} use a BERT-like model to initialize the lower layers of a generator, without finetuning.
Our work differs from these in that we attempt to directly generate from the pretrained model, rather than using it as a starting point to learn another model. 
%The advantage of our approach is that we can avoid expensive finetuning of these large models, with minimal performance loss.

\subsection{Gibbs sampling in the generalized sequence generation model}
\label{subsec:gibbs_sampling}

\paragraph{Gibbs sampling: uniform coordinate selection}

%We first investigate how we could frame Gibbs sampling in the proposed generalized sequence generation framework. 
To cast Gibbs sampling into our framework,
%Similar to the case of non-autoregressive neural sequence model \ref{sec:nonauto}, 
we first assume that the length prediction distribution $P(L|X)$ is estimated from training data, as is the case in the non-autoregressive neural sequence model.
In Gibbs sampling, we often uniformly select a new coordinate at random, which corresponds to 
% the following coordinate selection distribution:
% \begin{align}
%     \label{eq:coordinate_selection_gibbs}
    \mbox{$p(z_i^{t+1} = 1 | Y^{\leq t}, Z^{t}, X) = 1/L$}
% \end{align}
with the constraint that $\sum_{i=1}^L z_i^t = 1$. By using the conditional BERT distribution as a symbol replacement distribution, we end up with Gibbs sampling.

\paragraph{Adaptive Gibbs sampling: non-uniform coordinate selection}\label{par:heuristic}

Instead of selecting coordinates uniformly at random,
% of the uniform coordinate selection procedure in Eq.~\eqref{eq:coordinate_selection_gibbs}, 
we can base selections on the intermediate sequences. We propose a log-linear model with features $\phi_i$ based on the intermediate and coordinate sequences:
%we can use a log-linear model so that selection depends on intermediate sequences:
\resizebox{0.96\linewidth}{!}{
\begin{minipage}{\linewidth}
\begin{align}
    \label{eq:coordinate_selection_adaptivegibbs_general}
    p(z_i^{t+1} = 1 | Y^{\leq t}, Z^{t}, X) 
    \propto
    \exp \left\{
    \frac{1}{\tau} \sum_{i=1}^L \alpha_i \phi_i(Y^t, Z^{t}, X, i)
    \right\}
\end{align}
\end{minipage}
}
again with the constraint that $\sum_{i=1}^L z_i^t = 1$.
%similarly to Gibbs sampling. 
$\tau > 0$ is a temperature parameter controlling the sharpness of the coordinate selection distribution. A moderately high $\tau$ smooths the coordinate selection distribution and ensures that all the coordinates are replaced in the infinite limit of $T$, making it a valid Gibbs sampler~\citep{levine2006optimizing}.

We investigate three features $\phi_i$: % in this paper. 
(1) We compute how peaked the conditional distribution of each position is given the symbols in all the other positions by measuring its {\it negative entropy}:
% \begin{align*}
    % \label{eq:feature_entropy}
    $\phi_{\text{negent}}(Y^t, Z^{t}, X, i) = 
    -
    \mathcal{H}(y^{t+1}_{i} | y^t_{<i}, \left< \text{mask} \right>, y^t_{>i}, X).$
    % \sum_{v \in V} p(y^{t+1}_{i}=v | y^t_{<i}, \left< \text{mask} \right>, y^t_{>i}, X) \log p(y^{t+1}_{i}=v | y^t_{<i}, \left< \text{mask} \right>, y^t_{>i}, X).
% \end{align*}
In other words, we prefer a position $i$ if we know the change in $i$ has a high potential to alter the joint probability $p(Y|X) = p(y_1, y_2, ..., y_L|X)$.
(2) For each position $i$ we measure how unlikely the {\it current } symbol ($y^t_i$, not $y^{t+1}_i$) is under the {\it new} conditional distribution:
% \begin{align}
    % \label{eq:feature_logp}
    $\phi_{\text{logp}}(Y^t, Z^t, X, i) = 
    -\log p(y_i=y^t_i | y^t_{<i}, \left< \text{mask} \right>, y^t_{>i}, X)$.
% \end{align}
%It is not $y_i=y^{t+1}_i$ as in $\phi_{\text{negent}}$ or in Eq.~\eqref{eq:generalized_sequence_generation}~(b) but $y_i=y^t_i$. 
Intuitively, we prefer to replace a symbol if it is highly incompatible with the input variable and all the other symbols in the current sequence.
(3) We encode a {\it positional preference} that does not consider the content of intermediate sequences:
% \begin{align}
%     \label{eq:feature_pos}
    $\phi_{\text{pos}}(i) = 
    - \log (| t - i |+\epsilon)$,
% \end{align}
where $\epsilon > 0$ is a small constant scalar to prevent $\log 0$. This feature encodes our preference to generate from left to right if there is no information about 
% the content of 
the input variable nor of any intermediate sequences.

%By combining above three features \eqref{eq:feature_entropy}--\eqref{eq:feature_pos} together, we end up with the following log-linear model:
%\begin{align}
%    \label{eq:coordinate_selection_adaptivegibbs}
%    \log p(z_i^{t+1} = 1 | Y^t, Z^{t}, X) 
%    = 
%    \frac{1}{\tau} \left(
%    \sum_{i=1}^L \alpha_{\text{negent}} \phi_{\text{negent}}(Y^t, Z^{t}, X, i) + \alpha_{\text{logp}} \phi_{\text{logp}}(Y^t, Z^{t}, X, i) + \alpha_{\text{pos}} \phi_{\text{pos}}(i)
%    \right)
%    + C,
%\end{align}
%where $C$ is a constant.

Unlike the special cases of the proposed generalized model in \S\ref{sec:generalized_sequence_generation}, the coordinate at each generation step is selected based on the intermediate sequences, previous coordinate sequences, and the input variable. 
%This allows us to be flexible in selecting positions in the sequence according to the input variable and intermediate generated sequences. 
We mix the features using scalar coefficients $\alpha_{\text{negent}}$, $\alpha_{\text{logp}}$ and  $\alpha_{\text{pos}}$, which are selected or estimated to maximize a target quality measure on the validation set.

\paragraph{Adaptive Gibbs sampling: learned coordinate selection}\label{par:learned} 
% We can parameterize the coordination selection distribution, and learn parameters in order to select coordinates which maximize a reward function. 
% In this case, we refer to the coordinate selection distribution as a \textit{policy}, $\pi_{\theta}(z_i^{t+1}|s_t)$, where the state $s_t$ is $(Y^{\leq t},Z^t,X)$. A state transition, $s_{t+1}\sim p(\cdot|s_t,z_i^{t+1})$, consists of generating a symbol at position $z_i^{t+1}$.
% Given a scalar reward function $r(s_t,a_t,s_{t+1})$, the objective is to find parameters that maximize  expected future reward, with the expectation taken over the distribution of generations obtained using the policy for coordinate selection,
% \begin{align}
% \label{eq:rl-objective}
%     J(\theta)&= \sum_{t=0}^{T-1}\mathbb{E}_{\substack{z\sim \pi_{\theta}(\cdot|s_t)\\s_{t+1}\sim p(\cdot|s_t,z)}}\left[\sum_{t'=t}^{T-1}\gamma^{(t'-t)}r(s_{t'},z,s_{t'+1})\right],
% \end{align}
% where $\gamma\in (0,1]$ is a discount factor, and $s_0$ is an input $X$ along with an empty output and coordinate sequence. 
% We maximize this objective by estimating its gradient using 
% %For optimization, gradients of $J(\theta)$ are obtained using a gradient estimator such as 
% REINFORCE \cite{williams1992simple}.
% We discuss our choice of reward function, policy parameterization, and hyperparameters later in Section \ref{sec:experimental-settings}.

We learn a coordinate selection distribution that selects coordinates in order to maximize a reward function that we specify. 
In this case, we refer to the coordinate selection distribution as a 
\textit{policy}, $\pi_{\theta}(a_t|s_t)$, 
where a state $s_t$ is $(Y^{\leq t},Z^t,X)$, 
an action $a_t\in\{1,\ldots,L\}$ is a coordinate, 
so that $Z^{t+1}$ is 1 at position $a_t$ and 0 elsewhere, 
and $\pi_{\theta}$ is parameterized using a neural network. %
%\footnote{See Section 4 for details of the model used in the experiments. Our experiments include test-time constant-time decoding, where we select a set of actions $A_t$ and $Z^{t+1}$ is 1 at each $a\in A_t$.}
Beginning at a state $s_1\sim p(s_1)$ corresponding to an input $X$ 
%(e.g. sampled from a dataset) 
along with an empty coordinate and output sequence, we obtain a generation by repeatedly sampling a coordinate $a_t\sim \pi_{\theta}(\cdot|s_t)$ and transitioning to a new state for $T$ steps. 
Each transition, $s_{t+1}\sim p(\cdot|s_t,a_t)$, consists of generating a symbol at position $a_t$.
Given a scalar reward function $r(s_t,a_t,s_{t+1})$, the objective is to find a policy that maximizes expected reward, with the expectation taken over the distribution of generations obtained using the policy for coordinate selection,
\begin{align}
\label{eq:rl-objective}
    J(\theta)&= \mathbb{E}_{\tau\sim \pi_{\theta}(\tau)}\left[\sum_{t=1}^{T}\gamma^{t-1}r(s_{t},a_t,s_{t+1})\right],\\
    \pi_{\theta}(\tau) &=p(s_1)\prod_{t=1}^{T}\pi_{\theta}(a_t|s_t)p(s_{t+1}|a_t,s_t),
\end{align}
where $\tau =(s_1, a_1,s_2,\ldots,a_{T},s_{T+1})$, and $\gamma\in [0,1]$ is a discount factor (with $0^0=1$).
We maximize this objective by estimating its gradient using policy gradient methods \cite{williams1992simple}.
We discuss our choice of reward function, policy parameterization, and hyperparameters later in Section \ref{sec:experimental-settings}.

\subsection{Optimistic decoding and beam search from a masked language model}
\label{subsec:optimistic_decoding}

Based on the adaptive Gibbs sampler with the non-uniform and learned coordinate selection distributions % in paragraphs~\ref{par:heuristic} and ~\ref{par:learned} %Eq.~\eqref{eq:coordinate_selection_adaptivegibbs_general},
%and Eq.~\eqref{eq:coordinate_selection_adaptivegibbs}, 
we can now design an inference procedure to approximately find the most likely sequence $\argmax_{Y} p(Y|X)$ from the sequence distribution by exploiting the corresponding model of sequence generation. In doing so, a naive approach is to marginalize out the generation procedure $G$ using a Monte Carlo method:
$\argmax_{Y^T} \frac{1}{M} \sum_{G^{m}} p(Y^T|Y^{m, < T}, Z^{m, \leq T}, X)$
% \begin{align}
% \label{eq:exact_inference}
%    \mbox{$\argmax_{Y} p(Y|X)
%    = 
%    \argmax_{Y^T} \frac{1}{M} \sum_{G^{m}} p(Y^T|Y^{m, < T}, Z^{m, \leq T}, X)$},
% \end{align}
where $G^m$ is the $m$-th sample from the sequence generation model. This approach suffers from a high variance and non-deterministic behavior, and is less appropriate for practical use. 
We instead propose an optimistic decoding approach following equation~\eqref{eq:generalized_sequence_generation}:
\resizebox{.99\linewidth}{!}{
\begin{minipage}{\linewidth}
\begin{align}
\label{eq:proposed_inference}    
\underset{\substack{L, Y^1, \ldots, Y^T \\ Z^1, \ldots, Z^T}}{\argmax} \log p(L|X) + &\sum_{t=1}^T \sum_{i=1}^L \Big(\log p(z^{t+1}_i | Y^{\leq t}, Z^{t}, X)\\
&+ z^{t+1}_i \log p(y^{t+1}_i | Y^{\leq t}, X)\Big)
\nonumber
\end{align}
\end{minipage}
}
The proposed procedure is {\it optimistic} in that we consider a sequence generated by following the most likely generation path to be highly likely under the sequence distribution obtained by marginalizing out the generation path. This optimism in the criterion more readily admits a deterministic approximation scheme such as greedy and beam search, although it is as intractable to solve this problem as the original problem 
% in Eq.~\eqref{eq:exact_inference} 
which required marginalization of the generation path.

\paragraph{Length-conditioned beam search}

To solve this intractable optimization problem, we design a heuristic algorithm, called length-conditioned beam search. 
Intuitively, given a length $L$, this algorithm performs beam search over the coordinate and intermediate token sequences.
At each step $t$ of this iterative algorithm, we start from the hypothesis set $\mathcal{H}^{t-1}$ that contains $K$ generation hypotheses:
% \[
\mbox{$\mathcal{H}^{t-1} = \left\{
h^{t-1}_k = (
    (\hat{Y}_k^1, \ldots, \hat{Y}_k^{t-1}),
    (\hat{Z}_k^1, \ldots, \hat{Z}_k^{t-1})
)
\right\}_{k=1}^K$}.
% \]
Each generation hypothesis has a score:
\resizebox{.99\linewidth}{!}{
\begin{minipage}{\linewidth}
\begin{align*}
s(h^{t-1}_k) = \log p(L|X) + &\sum_{t'=1}^{t-1} \sum_{i=1}^L \Bigg(\log p(\hat{z}^{t'}_i | \hat{Y}^{< t'}_k, \hat{Z}^{t'-1}, X)\\
&+ \hat{z}^{t'}_i \log p(\hat{y}^{t'}_i | \hat{Y}^{\leq t}, X)\Bigg).
\end{align*}
\end{minipage}
}
%\[
%s(h^{t-1}_k) = &\log p(L|X) + \sum_{t'=1}^{t-1} \sum_{i=1}^L \log p(\hat{z}^{t'}_i | \hat{Y}^{< t'}_k, \hat{Z}^{t'-1}, X) + \hat{z}^{t'}_i \log p(\hat{y}^{t'}_i | \hat{Y}^{\leq t}, X).
%\]

For notational simplicity, we drop the time superscript $t$.
Each of the $K$ generation hypotheses is first expanded with $K'$ candidate positions according to the coordinate selection distribution: 
% \[
%\mbox{$\hat{h}_{k, k'} = h_k \| z_{k, k'}$},
% \]
%where $\cdot \| \cdot$ denotes concatenating $Z$ or $Y$ to the respective sequence in $h_k$ (in the case of the former, the sequence of $Z$s is one longer than the sequence of $Y$s), and
%is a shorthand notation for concatenation, and 
\resizebox{.99\linewidth}{!}{
\begin{minipage}{\linewidth}
\begin{align*}
\text{arg~top-$K'$}_{u \in \{1, \dots, L\} } 
\underbrace{s(h_k) + \log p(z_{k, u} = 1| \hat{Y}^{< t}, \hat{Z}^{t-1}, X)}_{
% =s(h_k \| z_{k, j})
=s(h_k \| \text{one-hot}(u) )}
\end{align*}
\end{minipage}
}
\iffalse
\begin{align*}
&\left\{ \hat{h}_{k,k'} \right\}_{k'=1}^{K'} =\\
&= \text{arg~top-$K'$}_{u \in \{1, \dots, L\} } 
\underbrace{s(h_k) + \log p(z_{k, u} = 1| \hat{Y}^{< t}, \hat{Z}^{t-1}, X)}_{
% =s(h_k \| z_{k, j})
=s(h_k \| \text{one-hot}(u) )}
\end{align*}
\fi
%\[
%\left\{ \hat{h}^{t}_{k,k'} \right\}_{k'=1}^{K'} = \text{argtop-$K'$}_{z \in \left\{0, 1 \right\}^L; \sum_{i=1}^L z_i =1 } 
%\underbrace{s(h^{t-1}_k) + \sum_{i=1}^L \log p(z^{t}_i | \hat{Y}^{< t}, \hat{Z}^{t-1}, X)}_{=s(h^{t-1}_k \| z)}
%\underbrace{s(h^{t-1}_k) + \sum_{i=1}^L \log p(z^{t}_i | \hat{Y}^{< t}, \hat{Z}^{t-1}, X)}_{=s(h^{t-1}_k \| z)}
%\]
so that we have $K \times K'$ candidates $\left\{ \hat{h}_{k,k'} \right\}$, where each candidate consists of a hypothesis $h_k$ with the position sequence extended by the selected position 
% $z_{k, k'}$ 
$u_{k,k'}$
and has a score 
% $s(h_{k} \| z_{k,k'})$.
$s(h_k \| \text{one-hot}(u_{k,k'}))$.\footnote{
$h_k \| \text{one-hot}(u_{k,k'})$ appends $\text{one-hot}(u_{k,k'})$ at the end of the sequence of the coordinate sequences in $h_{k}$
}
% $s(\hat{h}_{k,k'})$.
We then expand each candidate with the symbol replacement distribution:
%Each of these $K \times K'$ candidates $\left\{ \hat{h}_{k,k'} \right\}$ is then further expanded with symbol replacement:
% \[
%\mbox{$\hat{\hat{h}}_{k,k',k''}^{t-1} = \hat{h}_{k,k'}^{t-1} \| y^{t}_{k,k',k''}$},
% \]
%where
\resizebox{.99\linewidth}{!}{
\begin{minipage}{\linewidth}
\begin{align*}
&\text{arg~top-$K''$}_{v \in V}
\underbrace{s(h_k \| \text{one-hot}(u_{k,k'})) + \log p(y_{z_{k, k'}} = v | \hat{Y}^{\leq t}, X)}_{
% =s(h_{k, k'} \| y')
=s(h_{k, k'} \| (\hat{Y}^{t-1}_{<z_{k,k'}},v,\hat{Y}^{t-1}_{>z_{k,k'}}))
}.
\end{align*}
\end{minipage}
}
\iffalse
\begin{align*}
&\left\{ 
\hat{\hat{h}}_{k, k', k''}
\right\}_{k''=1}^{K''}
=\\
&=\text{arg~top-$K''$}_{v \in V}
\underbrace{s(h_k \| \text{one-hot}(u_{k,k'})) + \log p(y_{z_{k, k'}} = v | \hat{Y}^{\leq t}, X)}_{
% =s(h_{k, k'} \| y')
=s(h_{k, k'} \| (\hat{Y}^{t-1}_{<z_{k,k'}},v,\hat{Y}^{t-1}_{>z_{k,k'}}))
}.
\end{align*}
\fi
This results in $K \times K' \times K''$ candidates $\left\{\hat{\hat{h}}_{k, k', k''}\right\}$, each consisting of hypothesis $h_k$ with intermediate and coordinate sequence respectively extended by $v_{k,k',k''}$ and $u_{k,k'}$. Each hypothesis has a score 
$s(h_{k, k'} \| (\hat{Y}^{t-1}_{<z_{k,k'}},v_{k, k',k''},\hat{Y}^{t-1}_{>z_{k,k'}}))$,\footnote{
$h_{k, k'} \| (\hat{Y}^{t-1}_{<z_{k,k'}},v_{k, k',k''},\hat{Y}^{t-1}_{>z_{k,k'}})$ denotes
creating a new sequence from $\hat{Y}^{t-1}$ by replacing the \mbox{$z_{k, k'}$-th} symbol with $v_{k, k',k''}$, and appending this sequence to the intermediate sequences in $h_{k,k'}$.
} 
% $s(\hat{\hat{h}}_{k, k', k''})$, 
which we use to select $K$ candidates to form a new hypothesis set 
% \[
\mbox{$\mathcal{H}^t = \text{arg~top-$K$}_{h \in \left\{ \hat{\hat{h}}_{k,k',k''} \right\}_{k,k',k''}} s(h)$}.
% \]

After iterating for a predefined number $T$ of steps, the algorithm terminates with the final set of $K$ generation hypotheses. We then choose one of them according to a prespecified criterion, such as Eq.~\eqref{eq:proposed_inference}, and return the final symbol sequence $\hat{Y}^T$.

\begin{table*}[t]
    \centering
    \small
\begin{tabular}[t]{ccc||c|ccccc}
& & & 
Baseline
& 
\multicolumn{4}{c}{Decoding from an undirected sequence model}
\\
& $b$ & $T$ & 
\rotatebox{0}{Autoregressive} & 
\rotatebox{0}{Uniform} & 
\rotatebox{0}{Left2Right} &
\rotatebox{0}{Least2Most} &
\rotatebox{0}{Easy-First} &
\rotatebox{0}{Learned}
\\
\toprule
\multirow{4}{*}{\rotatebox{90}{En$\to$De}}
& $1$ & $L$ & 
25.33 &
21.01 &
24.27 &
23.08 &
23.73 &
24.10
\\
& $4$ & $L$ &
26.84 &
22.16 & 
25.15 & 
23.81 & 
24.13 &
24.87 % length penalty 1.0. 24.90 with length penalty 0.6.
\\
& $4$ & $L$\text{*} &
-- &
22.74 & 
\textbf{25.66} & 
24.42 & 
24.69 &
\textbf{25.28}
\\
& $1$ & $2L$ &
-- &
21.16 &
24.45 &
23.32 &
23.87 &
24.15
\\
& $4$ & $2L$ &
-- &
21.99 &
25.14 &
23.81 &
24.14 &
24.86 % length penalty 1.0
\\
\midrule
\multirow{4}{*}{\rotatebox{90}{De$\to$En}}
& $1$ & $L$ & 
29.83 &
26.01 & 
28.34 &
28.85 &
29.00 &
28.47
\\
& $4$ & $L$ &
30.92 &
27.07 &
29.52 & 
29.03 &
29.41 &
29.73  % length penalty 1.0. 29.54 with length penalty 0.6.
\\
& $4$ & $L$\text{*} &
-- &
28.07 &
\textbf{30.46} & 
29.84 &
30.32 &
\textbf{30.58}
\\
& $1$ & $2L$ &
-- &
26.24 &
28.64 &
28.60 &
29.12 &
28.45
\\
& $4$ & $2L$ &
-- &
26.98 &
29.50 &
29.02 &
29.41 &
29.71 % length penalty 1.0. 29.54 with length penalty 0.6.
\end{tabular}

\caption{Results (BLEU$\uparrow$) on WMT'14 En$\leftrightarrow$De translation using various decoding algorithms and different settings of beam search width ($b$) and number of iterations ($T$) as a function of sentence length ($L$). For each sentence we use $4$ most likely sentence lengths. \text{*} denotes rescoring generated hypotheses using autoregressive model instead of proposed model.}
    \label{tab:bleu_performance2}
    
    \vspace{-4mm}
\end{table*}    

\section{Experimental Settings}
\label{sec:experimental-settings}
\paragraph{Data and preprocessing}

We evaluate our framework on WMT'14 English-German translation.
%We evaluate our framework on WMT'14 En$\leftrightarrow$De translation, a standard benchmark for evaluating sequence-to-sequence models. 
%We choose to work with WMT'14 En$\leftrightarrow$De dataset, which consists of 4.5M pairs of parallel English$\leftrightarrow$German sentences and is used as standard benchmark for sequence-to-sequence tasks. 
The dataset consists of 4.5M parallel sentence pairs.
%, and we follow the widely used protocol for preprocessing this dataset. 
We preprocess this dataset by tokenizing each sentence using a script from Moses~\citep{koehn2007moses} and then segmenting each word into subword units using byte pair encoding~\citep{sennrich2016bpe} with a joint vocabulary of 60k tokens. We use newstest-2013 and newstest-2014 as validation and test sets respectively. 

% \subsection{Models and Learning}

\paragraph{Sequence models} 

We base our models off those of \citet{lample2019cross}.
% \footnote{ 
% \url{https://github.com/facebookresearch/XLM}
% }
Specifically, we use a Transformer \citep{vaswani2017attention} with 1024 hidden units, 6 layers, 8 heads, and Gaussian error linear units  \citep{hendrycks2016gelu}.
%We build our models using publicly available framework of cross-lingual masked language models~\citep{lample2019cross}.\footnote{ \url{https://github.com/facebookresearch/XLM}}  
We use a pretrained model\footnote{ \url{https://dl.fbaipublicfiles.com/XLM/mlm_ende_1024.pth}} trained using a masked language modeling objective \citep{lample2019cross} on 5M monolingual sentences from WMT NewsCrawl 2007-2008.
%We use a pretrained multi-lingual BERT Masked Language Model\footnote{
%\url{https://dl.fbaipublicfiles.com/XLM/mlm_ende_1024.pth}
%} 
%trained on large English and German monolingual corpora of WMT'14 proceedings (news-2007, news-2008). 
To distinguish between English and German sentences, a special language embedding is added as an additional input to the model. 

We adapt the pretrained model to translation by finetuning it with a masked translation objective \citep{lample2019cross}. We concatenate parallel English and German sentences,
%with their corresponding language tag from training set, 
mask out a subset of the tokens in either the English or German sentence, and predict the masked out tokens. We uniformly mask out $0-100\%$ tokens as in \citet{ghazvininejad2019constant}.
Training this way more closely matches the generation setting, where the model starts with an input sequence of all masks.

\paragraph{Baseline model} 
%\paragraph{Baseline model: an autoregressive transformer} 

We compare against a standard Transformer encoder-decoder autoregressive neural sequence model \citep{vaswani2017attention} trained for left-to-right generation and initialized with the same pretrained model. %\citep{lample2019cross,song2019mass}. 
We train a separate autoregressive model to translate an English sentence to a German sentence and vice versa, with the same hyperparameters as our model. 
%These models were also initialized using pretrained weights of masked language model from \citet{lample2019cross} and use same hyperparameters as in our model.

\paragraph{Training details}

We train the models using Adam \citep{kingma2014adam} with an inverse square root learning rate schedule, learning rate of $10^{-4}$, $\beta_1 = 0.9$, $\beta_2=0.98$, and dropout rate of $0.1$ \citep{srivastava2014dropout}. 
Our models are trained on $8$ GPUs with a batch size of $256$ sentences. 

% \subsection{Decoding strategies and scenarios}

\paragraph{Handcrafted decoding strategies}

\begin{table}[h!]
%\begin{wraptable}{R}{0.5\textwidth}
    %\vspace{-5mm}
	\small
    \centering
\begin{tabular}{c||c|cccc}
& & \multicolumn{4}{c}{\# of length candidates}
\\
& Gold & 1 & 2 & 3 & 4 
\\
\toprule
En$\to$De & 22.50 & 22.22 & 22.76 & 23.01 & \textbf{23.22}
\\
De$\to$En & 28.05 & 26.77 & 27.32 & 27.79 & \textbf{28.15} 
\end{tabular}

% \begin{tabular}[b]{llcc}
% \toprule
% & & \multicolumn{1}{c}{WMT'14 En$\rightarrow$De} & \multicolumn{1}{c}{WMT'14 De$\rightarrow$En} \\
%   \midrule
%   \midrule
%  \multirow{5}{*}{\rotatebox[origin=c]{90}{\scriptsize Our Model}}
%     & Gold Length & 22.39 & 28.05 \\
% 	& $L$\;=\;1 & 21.98 & 26.77\\
% 	& $L$\;=\;2 & 22.58 & 27.32 \\
% 	& $L$\;=\;3 & 22.92 & 27.79 \\
% 	& $L$\;=\;4 & 23.13 & 28.15 \\
%     \bottomrule
% \end{tabular}
% \vspace{-2mm}
\caption{Effect of the number of length candidates considered during decoding on BLEU, measured on the validation set (newstest-2013) using the {\bf easy-first} strategy.}
	\label{tab:bleu_length}
% \end{table*}  
%\vspace{-4mm}
%\end{wraptable}
\end{table}

We design four generation strategies for the masked translation model based on the log-linear coordinate selection distribution in \S\ref{eq:coordinate_selection_adaptivegibbs_general}:
\begin{enumerate}
\itemsep 0em
    \item {\bf Uniform}: $\tau \to \infty$, i.e., sample a position uniformly at random without replacement 
    % (eqivalent to the uniform coordinate selection in Gibbs sampling)
    \item {\bf Left2Right}: $\alpha_{\text{negent}}=0$, $\alpha_{\text{logp}}=0$, $\alpha_{\text{pos}}=1$
    \item {\bf Least2Most}~\citep{ghazvininejad2019constant}: $\alpha_{\text{negent}}=0$, $\alpha_{\text{logp}}=1$, $\alpha_{\text{pos}}=0$
    \item {\bf Easy-First}: $\alpha_{\text{negent}}=1$, $\alpha_{\text{logp}}=1$,\footnote{
    We set $\alpha_{\text{logp}}=0.9$ for De$\to$En based on the validation set performance.
    } 
    $\alpha_{\text{pos}}=0$
\end{enumerate}

We use beam search described in \S\ref{subsec:optimistic_decoding} with $K'$ fixed to $1$, i.e., we consider only one possible position for replacing a symbol per hypothesis each time of generation. We vary $K = K''$ between $1$ (greedy) and $4$. For each source sentence, we consider four length candidates according to the length distribution estimated from the training pairs, 
%as we observe as high quality translations using four length candidates as those decoded using the ground-truth length 
based on early experiments showing that using only four length candidates performs as well as using the ground-truth length
(see Table~\ref{tab:bleu_length}). 
%Among the four decoded translations of varying lengths, 
Given the four candidate translations,
we choose the best one according to the pseudo log-probability of the final sequence~\citep{wang2019bert}. Additionally, we experimented with choosing best translation according to log-probability of the final sequence calculated by an autoregressive neural sequence model.

\paragraph{Learned decoding strategies}

We train a parameterized coordinate selection policy to maximize expected reward (Eq. \ref{eq:rl-objective}). As the reward function, we use the change in edit distance from the reference,
\begin{align*}
    r(s_t,a_t,s_{t+1})=(d_{\text{edit}}(Y^{\leq t},Y) - d_{\text{edit}}(Y^{\leq{t+1}},Y)),
\end{align*} 
where $s_t$ is $(Y^{\leq t},Z^t,X)$. The policy is parameterized as,
\begin{align*}
    \pi_{\theta}(a_t|s_t)=\text{softmax}\left(f_{\theta}(h_1, \bar{h}),\ldots,f_{\theta}(h_L, \bar{h})\right),
\end{align*}
where $h_i\in\mathbb{R}^{1024}$ is the masked language model's output vector for position $i$, and $\bar{h}\in\mathbb{R}^{1024}$ is a history of the previous $k$ selected positions, $\bar{h}=\frac{1}{k}\sum_{j=1}^k (\text{emb}_{\theta}(j)+h^{j}_{a_j})$. 
We use a 2-layer MLP for $f_{\theta}$ which concatenates its inputs and has hidden dimension of size 1024.

Policies are trained with linear time decoding ($T=L$), with positions sampled from the current policy, and symbols selected greedily. 
At each training iteration we sample a batch of generations, add the samples to a FIFO buffer, then perform  gradient updates on batches sampled from the buffer. 
We use proximal policy optimization (PPO), specifically the clipped surrogate objective from \citet{schulman2017proximal} with a learned value function $V_{\theta}(s_t)$ to compute advantages. 
This objective resulted in stable training compared to initial experiments with REINFORCE \cite{williams1992simple}. The value function is a 1-layer MLP, $V_{\theta}(\frac{1}{L}\sum_{i=1}^L(h_i,\bar{h}))$. 

Training hyperparameters were selected based on validation BLEU in an initial grid search of $\text{generation batch size}\in \{4,16\}$ (sequences), $\text{FIFO buffer size}\in \{1k,10k\}$ (timesteps), and $\text{update batch size}\in \{32,128\}$ (timesteps). Our final model was then selected based on validation BLEU with a grid search on $\text{discount }\gamma\in\{0.1,0.9,0.99\}$ and $\text{history }k\in\{0,20,50\}$ for each language pair, resulting in a discount $\gamma$ of $0.9$ for both pairs, and history sizes of $0$ for De$\rightarrow$En and $50$ for En$\rightarrow$De.

% We consider linear-time and constant-time settings during decoding with our model.
% In the linear-time setting, we use both greedy and beam search decoding ($b = 4$). We also investigate doing a second decoding pass over generated sequence.
% In the constant-time setting, we select the number of positions to predict at each time step as a function of the sequence length $L$ and budget size $T$. We consider two schedules: linearly annealing the number of selected positions from $L$ to $1$~\citep{ghazvininejad2019constant} and fixing the number of selected positions to $\lceil L/T \rceil$. 

% {\color{red} KC: BEGIN}
\paragraph{Decoding scenarios}

We consider two decoding scenarios: linear-time and constant-time decoding. In the linear-time scenario, the number of decoding iterations $T$ grows linearly w.r.t. the length of a target sequence $L$. We test setting $T$ to $L$ and $2L$. In the constant-time scenario, the number of iterations is constant w.r.t. the length of a translation, i.e., $T=O(1)$. At the $t$-th iteration of generation, we replace $o_t$-many symbols, where $o_t$ is either a constant $\lceil L/T \rceil$ or linearly anneals from $L$ to $1$ ($L\to 1$) as done by \citet{ghazvininejad2019constant}. 

\section{Linear-Time Decoding: Result and Analysis}

\begin{figure*}[t]
    \centering
    \input{figures/fig-order.tex}
    \label{fig:gen-order}
    %\vspace{-5mm}
\end{figure*}

\paragraph{Main findings} 

We present translation quality measured by BLEU \citep{papineni2002bleu} in Table~\ref{tab:bleu_performance2}. We identify a number of important trends. 
(1) The deterministic coordinate selection strategies ({\bf left2right}, {\bf least2most}, {\bf easy-first} and {\bf learned}) significantly outperform selecting coordinates uniformly at random, by up to 3 BLEU in both directions. Deterministic coordinate selection strategies produce generations that not only have higher BLEU compared to uniform coordinate selection, but are also more likely according to the model as shown in Figures 1-2 in Appendix. The success of these relatively simple handcrafted and learned coordinate selection strategies suggest avenues for further improvement for generation from undirected sequence models.
%(1) The deterministic generation strategies with non-uniform coordinate selection ({\bf left2right}, {\bf least2most} and {\bf easy-first}) clearly outperform the naive variant ({\bf uniform}) by significant margins (up to 3 BLEU in both directions.) This success with a relatively simple and hand-tuned log-linear coordinate selection strategy implies that there is more room for improvement when it comes to generation from undirected sequence models. 
(2) The proposed beam search algorithm for undirected sequence models provides an improvement of about 1 BLEU over greedy search, confirming the utility of the proposed framework as a way to move decoding techniques across different paradigms of sequence modeling. 
(3) Rescoring generated translations 
%using beam search algorithm 
with an autoregressive model adds about 1 BLEU across all coordinate selection strategies.
Rescoring adds minimal overhead as it is run in parallel since the left-to-right constraint is enforced by masking out future tokens.
(4) Different generation strategies result in translations of varying qualities depending on the setting. {\bf Learned} and {\bf left2right} were consistently the best performing among all generation strategies. On English-German translation, {\bf left2right} is the best performing strategy slightly outperforming the {\bf learned} strategy, achieving 25.66 BLEU. On German-English translation, {\bf learned} is the best performing strategy, slightly outperforming the {\bf left2right} strategy while achieving 30.58 BLEU.
(5) We see little improvement in refining a sequence beyond the first pass.
(6) Lastly, the masked translation model is competitive with the state of the art neural autoregressive model, with a difference of less than 1 BLEU score in performance. We hypothesize that a difference between train and test settings causes a slight performance difference of the masked translation model compared to the conventional autoregressive model. In the standard autoregressive case, the model is explicitly trained to generate in left-to-right order, which matches the test time usage.  By randomly selecting tokens to mask during training, our undirected sequence model is trained to follow all possible generation orders and to use context from both directions, which is not available when generating left-to-right at test time.

\paragraph{Adaptive generation order}

\begin{table*}[t]
% 	\small
    \centering \small
\begin{tabular}[t]{cc||cccccc}
$T$ & $o_t$ & 
\rotatebox{0}{Uniform} & 
\rotatebox{0}{Left2Right} &
\rotatebox{0}{Least2Most} &
% \rotatebox{10}{Most2Least} &
\rotatebox{0}{Easy-First} &
\rotatebox{0}{Hard-First} &
\rotatebox{0}{Learned}
\\
\toprule
$10$ & $L\to 1$ & 
22.38 &
22.38 &
27.14 &
% 19.21 &
22.21 &
26.66 &
12.70
\\
$10$ & $L\to 1$\text{*} & 
23.64 &
23.64 &
\textbf{28.63} &
23.79 &
\textbf{28.46} &
13.18
\\
$10$ & $\lceil L/T \rceil$ &
22.43 &
21.92 &
24.69 &
% 0.09 &
25.16 &
23.46 &
26.47  % 26.91 w/ ar-rescoring 16
\\
\midrule
$20$ & $L\to 1$ & 
26.01 &
26.01 &
28.54 &
% 19.21 &
22.24 &
28.32 &
12.85
\\
$20$ & $L\to 1$\text{*} & 
27.28 &
27.28 &
\textbf{30.13} &
24.55 &
\textbf{29.82} &
13.19 \\
$20$ & $\lceil L/T \rceil$ &
24.69 &
25.94 &
27.01 &
% 0.00 &
27.49 &
25.56 &
27.82  % 28.54 w/ ar-rescoring 16
\\
\end{tabular}

\caption{Constant-time machine translation on WMT'14 De$\rightarrow$En with different settings of the budget ($T$) and number of tokens predicted each iteration ($o_t$). \text{*} denotes rescoring generated hypotheses using autoregressive model instead of proposed model.
}

\vspace{-4mm}
% $L \rightarrow 1$ indicates a linear decaying schedule of number of tokens predicted. $\lceil L/T \rceil$ indicates a constant number of tokens that is being predicted. For each sentence we use $4$ most likely sentence lengths.}
	\label{tab:faster_decoding}
    %\vspace{-6mm}
    % \vspace{-3mm}
\end{table*} 
The {\bf least2most}, {\bf easy-first}, and {\bf learned} generation strategies automatically adapt the generation order based on the intermediate sequences generated. 
We investigate the resulting generation orders on the development set by presenting each as a 10-dim vector (downsampling as necessary), where each element corresponds to the selected position in the target sequence normalized by sequence length. We cluster these sequences with $k$-means clustering and visualize the clusters centers as curves with thickness proportional to the number of sequences in the cluster in Fig.~\ref{fig:gen-order}.  

The visualization reveals that many sequences are generated monotonically, either left-to-right or right-to-left (see, e.g., {\color{green} green}, {\color{violet} purple} and {\color{orange} orange} clusters in {\bf easy-first}, De$\to$En, and {\color{orange} orange}, {\color{blue} blue}, and {\color{red} red} clusters in {\bf learned}, En$\to$De).
%{\color{green} green}, {\color{orange} orange}, and {\color{red} red} clusters in {\bf least2most}, En$\to$De).
%These orders are particularly common when using the learned policy.
For the \textbf{easy-first} and \textbf{least2most} strategies, we additionally identify clusters of sequences that are generated from outside in (e.g., {\color{blue} blue} and {\color{red} red} clusters in {\bf easy-first}, De$\to$En, and {\color{red} red} and {\color{violet} purple} clusters in {\bf least2most}, En$\to$De). %The outside in patterns are less frequent with the \textbf{learned} policy.

On De$\to$En, in roughly 75\% of the generations, the \textbf{learned} policy either generated from left-to-right ({\color{orange}orange}) or generated the final token, typically punctuation, followed by left-to-right generation ({\color{green}green}). 
In the remaining 25\% of generations, the \textbf{learned} policy generates with variations of an outside-in strategy ({\color{red}red}, {\color{blue}blue}, {\color{purple}purple}).
%, either beginning from the left or right end of the sequence.
See Appendix Figures 7-9 for examples.
On En$\to$De, the \textbf{learned} policy has a higher rate of left-to-right generation, with roughly 85\% of generations using a left-to-right variation ({\color{blue}blue}, {\color{orange}orange}). 
These variations are however typically not strictly monotonic; the learned policy usually starts with the final token, and often skips tokens in the left-to-right order before generating them at a later time.
We hypothesize that the learned policy tends towards variations of left-to-right since (a) left-to-right may be an easy strategy to learn, yet (b) left-to-right achieves reasonable performance.

In general, we explain the tendency towards either monotonic or outside-in generation
%in the \textbf{easy-first} and \textbf{least2most} heuristc strategies
by the availability of contextual evidence, or lack thereof. At the beginning of generation, the only two non-mask symbols are the beginning and end of sentence symbols, making it easier to predict a symbol at the beginning or end of the sentence. As more symbols are filled near the boundaries, more evidence is accumulated for the decoding strategy to accurately predict symbols near the center. This process manifests either as monotonic or outside-in generation.
%\alert{sentence about why RL method recovers L->R / R->L}.
%We present sample sequences generated using these strategies in the appendix.

\section{Constant-Time Decoding: Result and Analysis}

The trends in constant-time decoding noticeably differ from those in linear-time decoding. First, the {\bf left2right} strategy performs comparably worse compared to the best performing strategies in constant-time decoding. The performance gap is wider (up to 4.8 BLEU) with a tighter budget ($T=10$). Second, the {\bf learned} coordinate selection strategy performs best when generating $\lceil L/T \rceil$ symbols every iteration, despite only being trained with linear-time decoding, but performs significantly worse when annealing the number of generated symbols from $L$ to $1$. This could be explained by the fact that the learned policy was never trained to refine predicted symbols, which is the case in $L \to 1$ constant-time decoding. Third, {\bf easy-first} is the second-best performing strategy in the $\lceil L/T \rceil$ setting, but similarly to the {\bf learned} strategy it performs worse in the $L \to 1$ setting. This may be because in the $L \to 1$ setting it is preferable to first generate hard-to-predict symbols and have multiple attempts at refining them, rather than predicting hard tokens at the end of generation process and not getting an opportunity to refine them, as is done in {\bf easy-first} scenario. To verify this hypothesis, we test a {\bf hard-first} strategy where we flip the signs of the coefficients of {\bf easy-first}  in the log-linear model. This new {\bf hard-first} strategy works on par with {\bf least2most}, again confirming that decoding strategies must be selected based on the target tasks and decoding setting. 

%Perhaps most importantly,
With a fixed budget of $T=20$, linearly annealing $o_t$ from $L$ to $1$, and {\bf least2most} decoding, constant-time translation can achieve translation quality comparable to linear-time translation with the same model (30.13 vs. 30.58), and to beam-search translations using the strong neural autoregressive model (30.13 vs 30.92).
%Perhaps most importantly, constant-time translation can achieve translation quality within 1 BLEU of comparable linear-time translation (28.54 vs. 29.52) with a fixed budget of $20$ and the annealing schedule with the right choice of decoding strategy ({\bf left2right}). 
Even with a tighter budget of $10$ iterations (less than half the average sentence length), constant-time translation loses only 1.8 BLEU points (28.63 vs. 30.58), which confirms the finding by \citet{ghazvininejad2019constant} and offers new opportunities in advancing constant-time machine translation systems. Compared to other constant-time machine translation approaches, our model outperforms many recent approaches by \citet{gu2018nonauto,lee2018deterministic,wang2019nonautoaux,ma2019flowseq}, while being comparable to \citet{ghazvininejad2019constant,shu2019lvnar}. We present full table comparing performance of various constant-time decoding approaches in Table 1 in Appendix.

\section{Conclusion}

We present a generalized framework of neural sequence generation that unifies decoding in directed and undirected neural sequence models. Under this framework, we separate position selection and symbol replacement, allowing us to apply a diverse set of generation algorithms, inspired by those for directed neural sequence models, to undirected models such as BERT and its translation variant. 

We evaluate these generation strategies on WMT'14 En-De machine translation using a recently proposed masked translation model. 
Our experiments reveal that undirected neural sequence models achieve performance comparable to conventional, state-of-the-art autoregressive models, given an appropriate choice of decoding strategy. 
%Our experiments reveal that undirected neural sequence models are almost, but not completely, competitive with conventional, state-of-the-art autoregressive models (within 1-2 BLEU) with an appropriate choice of decoding strategy. 
We further show that constant-time translation in these models performs similar to linear-time translation by using one of the proposed generation strategies. Analysis of the generation order automatically determined by these adaptive decoding strategies reveals that most sequences are generated either monotonically or outside-in.

%There are two limitations in this work. 
We only apply our framework to the problem of sequence generation. As one extension of our work, we could also apply it to other structured data such as grids (for e.g. images) and arbitrary graphs. Overall, we hope that our generalized framework opens new avenues in developing and understanding generation algorithms for a variety of settings.

\bibliography{main}
\bibliographystyle{icml2020}
\clearpage
\appendix
\section{Comparison with other non-autoregressive neural machine translation approaches}\label{ax:nonautoregressive}
We present the comparison of results of our approach with other constant-time machine translation approaches in Table~\ref{tab:contant_time_comparison}. Our model is most similar to conditional model by ~\cite{ghazvininejad2019constant}. However, there are differences in both model and training hyperparameters between our work and work by \cite{ghazvininejad2019constant}. We use smaller Transformer model with 1024 hidden units vs 2048 units in \citep{ghazvininejad2019constant}. We also train the model with more than twice smaller batch size since we use 8 GPUs on DGX-1 machine and \cite{ghazvininejad2019constant} use 16 GPUs on two DGX-1 machine with float16 precision. Finally we don’t average best 5 checkpoints and don’t use label smoothing for our model.

\section{Non-monotonic neural sequence models}

The proposed generalized framework subsumes recently proposed variants of non-monotonic generation~\citep{welleck2019nonmonotonic,stern2019inserttransformer,gu2019insertiondecoding}. Unlike the other special cases described above, these non-monotonic generation approaches learn not only the symbol replacement distribution but also the coordinate selection distribution, and implicitly the length distribution, from data. Because the length of a sequence is often not decided in advance, the intermediate coordinate sequence $Z^t$ and the coordinate selection distribution are reparameterized to work with relative coordinates rather than absolute coordinates. We do not go into details of these recent algorithms, but we emphasize that all these approaches are special cases of the proposed framework, which further suggests other variants of non-monotonic generation. 

\section{Energy evolution over generation steps}
\label{appendix:energy}

While the results in Table 1 in paper indicate that our decoding algorithms find better generations in terms of BLEU relative to uniform decoding, we verify that the algorithms produce generations that are more likely according to the model.
We do so by computing the energy (negative logit) of the sequence of intermediate sentences generated while using an algorithm, and comparing to the average energy of intermediate sentences generated by picking positions uniformly at random.
We plot this energy difference over decoding in Figure~\ref{fig:ef-lr-energy}. We additionally plot the evolution of energy of the sequence by different position selection algorithms throughout generation process in Figure~\ref{fig:ef-lr-energy2}.
Overall, we find that left-to-right, least-to-most, and easy-first do find sentences that are lower energy than the uniform baseline over the entire decoding process.
Easy-first produces sentences with the lowest energy, followed by least-to-most, and then left-to-right. 

\section{Sample sequences and their generation orders}
\label{appendix:samples}

We present sample decoding processes on De$\rightarrow$En with $b=1, T=L$ using the \textbf{easy-first} decoding algorithm in Figures~\ref{fig:gen-order-ex-1}, \ref{fig:gen-order-ex-2}, \ref{fig:gen-order-ex-3}, and \ref{fig:gen-order-ex-4}, and the \textbf{learned} decoding algorithm in Figures~\ref{fig:learned-gen-order-ex-1}, \ref{fig:learned-gen-order-ex-2}, and \ref{fig:learned-gen-order-ex-0}.
For easy-first decoding, we highlight examples decoding in right-to-left-to-right-to-left order, outside-in, left-to-right, and right-to-left orders, which respectively correspond to the {\color{orange} orange}, {\color{purple} purple}, {\color{red} red}, and {\color{blue} blue} clusters from Figure 1 in the main paper. 
For learned decoding, we highlight examples with right-to-left-to-right, outside-in, and left-to-right orders, corresponding to the {\color{blue} blue}, {\color{red} red}, and {\color{green} green} clusters.
The examples demonstrate the ability of the coordinate selection strategies to adapt the generation order based on the intermediate sequences generated. Even in the cases of largely monotonic generation order (left-to-right and right-to-left), each algorithm has the capacity to make small changes to the generation order as needed.

\begin{table*}[t!]
    \centering
    \centering \small
\begin{tabular}{lcc}
\toprule
 & \multicolumn{2}{c}{\textbf{WMT2014}} \\
\textbf{ Models} & \textbf{EN-DE} & \textbf{DE-EN} \\
\midrule\midrule
AR Transformer-base \citep{vaswani2017attention} & 27.30 & -- \\
\midrule\midrule
AR \citep{gu2018nonauto} & 23.4 & -- \\
NAR (+Distill +FT +NPD S=100) & 21.61 & -- \\
\midrule\midrule
AR \citep{lee2018deterministic} & 24.57 & 28.47 \\
Adaptive NAR Model & 16.56 & -- \\ 
Adaptive NAR Model (+Distill) & 21.54 & 25.43 \\
\midrule\midrule
AR \citep{wang2019nonautoaux} & 27.3 & 31.29 \\
NAT-REG (+Distill) & 20.65 & 24.77 \\ 
NAT-REG (+Distill +AR rescoring) & 24.61 & 28.90 \\
\midrule\midrule
AR \citep{ghazvininejad2019constant} & 27.74 & 31.09 \\
CMLM with 4 iterations & 22.25 & -- \\
CMLM with 4 iterations (+Distill) & 25.94 & 29.90 \\
CMLM with 10 iterations & 24.61 & -- \\
CMLM with 10 iterations (+Distill) & 27.03 & 30.53 \\
\midrule\midrule
AR \citep{shu2019lvnar} & 26.1 & -- \\
Latent-Variable NAR & 11.8 & -- \\
Latent-Variable NAR (+Distill) & 22.2 & -- \\
Latent-Variable NAR (+Distill +AR Rescoring) & 25.1 & -- \\
\midrule\midrule
AR \citep{ma2019flowseq} & 27.16 & 31.44 \\
FlowSeq-base (+NPD n = 30) & 21.15 & 26.04 \\
FlowSeq-base (+Distill +NPD n = 30) & 23.48 & 28.40 \\
\midrule\midrule
AR (ours) & 26.84 & 30.92 \\
Contant-time 10 iterations & 21.98 & 27.14 \\
Contant-time 10 iterations (+AR Rescoring) & 24.53 & 28.63 \\
Contant-time 20 iterations & 23.92 & 28.54 \\
Contant-time 20 iterations (+AR Rescoring) & 25.69 & 30.13 \\
\bottomrule
\end{tabular}
\caption{BLEU scores on WMT'14 En$\rightarrow$De and De$\rightarrow$En datasets showing performance of various constant-time machine translation approaches. Each block shows the performance of autoregressive model baseline with their proposed approach. AR denotes autoregressive model. Distill denotes distillation. AR rescoring denotes rescoring of samples with autoregressive model. FT denotes fertility. NPD denotes noisy parallel decoding followed by rescoring with autoregressive model. }
\label{tab:contant_time_comparison}
    \label{tab:constant_time_comparison}
\end{table*}

\begin{figure*}[t!]
    \centering
    \input{figures/fig-energy.tex}
    \label{fig:ef-lr-energy}
\end{figure*}

\begin{figure*}[t!]
    \centering
    \input{figures/fig-energy2.tex}
    \label{fig:ef-lr-energy2}
\end{figure*}

\begin{figure*}[t!]
    \centering
    \centering
\footnotesize
\begin{tabular}{cc}
%\begin{tabular}{cccccccccccccccccccc}
\toprule
Iteration & {\color{orange} Right-to-Left-to-Right-to-Left} \\
\midrule
(source) & Würde es mir je gelingen , an der Universität Oxford ein normales Leben zu führen ? \\
1 &  \_ \_ \_ \_ \_ \_ \_ \_ \_ \_ \_ \_ \_ \_ \_ \textbf{?} \\ 
2 &  \_ \_ \_ \_ \_ \_ \_ \_ \_ \_ \_ \_ \_ \_ \_ \textbf{Oxford} ? \\ 
3 &  \_ \_ \textbf{ever} \_ \_ \_ \_ \_ \_ \_ \_ \_ \_ \_ Oxford ? \\ 
4 &  \_ \textbf{I} ever \_ \_ \_ \_ \_ \_ \_ \_ \_ \_ \_ Oxford ? \\ 
5 &  \_ I ever \_ \_ \_ \_ \_ \_ \_ \_ \_ \_ \textbf{of} Oxford ? \\ 
6 &  \textbf{Would} I ever \_ \_ \_ \_ \_ \_ \_ \_ \_ \_ of Oxford ? \\ 
7 &  Would I ever \_ \_ \_ \_ \_ \textbf{normal} \_ \_ \_ \_ of Oxford ? \\ 
8 & Would I ever \_ \_ \_ \_ \_ normal \_ \textbf{at} \_ \_ of Oxford ? \\
9 & Would I ever \_ \_ \_ \_ \_ normal \_ at \textbf{the} \_ of Oxford ? \\
10 & Would I ever \_ \_ \_ \_ \_ normal \_ at the \textbf{University} of Oxford ? \\
11 & Would I ever \_ \_ \_ \_ \_ normal \textbf{life} at the University of Oxford ? \\
12 & Would I ever \_ \_ \_ \textbf{live} \_ normal life at the University of Oxford ? \\
13 & Would I ever \_ \_ \_ live \textbf{a} normal life at the University of Oxford ? \\
14 & Would I ever \_ \textbf{able} \_ live a normal life at the University of Oxford ? \\
15 & Would I ever \textbf{be} able \_ live a normal life at the University of Oxford ? \\
16 & Would I ever be able \textbf{to} live a normal life at the University of Oxford ? \\
(target) & Would I ever be able to lead a normal life at Oxford ? \\

\bottomrule
\end{tabular}

\caption{Example sentences generated following an {\color{orange} right-to-left-to-right-to-left} generation order, using the \textbf{easy-first} decoding algorithm on De$\rightarrow$En.}
    \label{fig:gen-order-ex-1}
\end{figure*}

\begin{figure*}[t!]
    \centering
    \centering
\footnotesize
\begin{tabular}{cc}
%\begin{tabular}{cccccccccccccccccccc}
\toprule
Iteration & {\color{violet} Outside-In} \\
\midrule
(source) & Doch ohne zivilgesellschaftliche Organisationen könne eine Demokratie nicht funktionieren . \\
1 &  \_ \_ \_ \_ \_ \_ \_ \_ \_ \_ \textbf{.} \\ 
2 &  \_ \_ \_ \_ \_ \_ \_ \_ \textbf{cannot} \_ . \\ 
3 &  \_ \_ \_ \_ \_ \_ \_ \textbf{democracy} cannot \_ . \\ 
4 &  \_ \textbf{without} \_ \_ \_ \_ \_ democracy cannot \_ . \\ 
5 &  \_ without \_ \_ \_ \_ \_ democracy cannot \textbf{work} . \\ 
6 &  \textbf{But} without \_ \_ \_ \_ \_ democracy cannot work . \\ 
7 &  But without \_ \_ \_ \_ \textbf{a} democracy cannot work . \\ 
8 &  But without \_ \textbf{society} \_ \_ a democracy cannot work . \\ 
9 &  But without \_ society \_ \textbf{,} a democracy cannot work . \\ 
10 & But without \textbf{civil} society \_ , a democracy cannot work . \\ 
11 & But without civil society \textbf{organisations} , a democracy cannot work . \\ 
(target) & Yet without civil society organisations , a democracy cannot function . \\

\bottomrule
\end{tabular}

\caption{Example sentences generated following an {\color{violet} outside-in} generation order, using the \textbf{easy-first} decoding algorithm on De$\rightarrow$En.}
    \label{fig:gen-order-ex-2}
\end{figure*}

\begin{figure*}[t!]
    \centering
    \centering
\footnotesize
\begin{tabular}{cc}
%\begin{tabular}{cccccccccccccccccccc}
\toprule
Iteration & {\color{red} Left-to-Right} \\
\midrule
(source) & Denken Sie , dass die Medien zu viel vom PSG erwarten ? \\
1 &  \_ \_ \_ \_ \_ \_ \_ \_ \_ \_ \_ \textbf{?} \\ 
2 & \textbf{Do} \_ \_ \_ \_ \_ \_ \_ \_ \_ \_ ? \\
3 &  Do \textbf{you} \_ \_ \_ \_ \_ \_ \_ \_ \_ ? \\
4 &  Do you \textbf{think} \_ \_ \_ \_ \_ \_ \_ \_ ? \\
5 &  Do you think \_ \_ \_ \_ \_ \_ \textbf{PS} \_ ? \\
6 &  Do you think \_ \_ \_ \_ \_ \_ PS \textbf{@G} ? \\
7 &  Do you think \_ \textbf{media} \_ \_ \_ \_ PS @G ? \\
8 &  Do you think \textbf{the} media \_ \_ \_ \_ PS @G ? \\
9 &  Do you think the media \textbf{expect} \_ \_ \_ PS @G ? \\
10 & Do you think the media expect \_ \textbf{much} \_ PS @G ? \\
11 & Do you think the media expect \textbf{too} much \_ PS @G ? \\
12 & Do you think the media expect too much \textbf{of} PS @G ? \\

(target) & Do you think the media expect too much of PS @G ? \\

\bottomrule
\end{tabular}

\caption{Example sentences generated following an {\color{red} left-to-right} generation order, using the \textbf{easy-first} decoding algorithm on De$\rightarrow$En.}
    \label{fig:gen-order-ex-3}
\end{figure*}

\begin{figure*}[t!]
    \centering
    \centering
\footnotesize
\begin{tabular}{cc}
%\begin{tabular}{cccccccccccccccccccc}
\toprule
Iteration & {\color{blue} Right-to-Left} \\
\midrule
(source) & Ein weiterer Streitpunkt : die Befugnisse der Armee . \\
1 &  \_ \_ \_ \_ \_ \_ \_ \_ \_ \_ \textbf{.} \\ 
2 &  \_ \_ \_ \_ \_ \_ \_ \_ \_ \textbf{army} . \\ 
3 &  \_ \_ \_ \_ \_ \_ \_ \textbf{of} \_ army . \\ 
4 &  \_ \_ \_ \_ \_ \_ \_ of \textbf{the} army . \\ 
5 &  \_ \_ \_ \_ \_ \_ \textbf{powers} of the army . \\ 
6 &  \_ \_ \_ \_ \_ \textbf{the} powers of the army . \\ 
7 &  \_ \_ \_ \_ \textbf{:} the powers of the army . \\ 
8 &  \_ \_ \textbf{point} : the powers of the army . \\ 
9 &  \_ \textbf{contentious} point : the powers of the army . \\ 
10 & \textbf{Another} contentious point : the powers of the army . \\
(target) & Another issue : the powers conferred on the army . \\

\bottomrule
\end{tabular}

\caption{Example sentences generated following an {\color{blue} right-to-left} generation order, using the \textbf{easy-first} decoding algorithm on De$\rightarrow$En.}
    \label{fig:gen-order-ex-4}
\end{figure*}

\begin{figure*}[t!]
    \centering
    \centering
\footnotesize
\begin{tabular}{cc}
%\begin{tabular}{cccccccccccccccccccc}
\toprule
Iteration & {\color{blue} Right-to-Left-to-Right} \\
\midrule
(source) & Die Aktien von Flight Centre stiegen gestern um 3 Cent auf 38,20 Dollar .\\
1 & \_ \_ \_ \_ \_ \_ \_ \_ \_ \_ \_ \_ \_ \textbf{.} \\
2 & \_ \_ \_ \_ \_ \_ \_ \_ \_ \_ \_ \_ \textbf{20} . \\
3 & \_ \_ \_ \_ \_ \_ \_ \_ \_ \_ \_ \textbf{38.} 20 . \\
4 & \_ \_ \_ \_ \_ \_ \_ \_ \_ \_ \textbf{\$} 38. 20 . \\
5 & \_ \_ \_ \_ \_ \_ \_ \_ \_ \textbf{to} \$ 38. 20 . \\
6 & \textbf{Flight} \_ \_ \_ \_ \_ \_ \_ \_ to \$ 38. 20 . \\
7 & Flight \textbf{Centre} \_ \_ \_ \_ \_ \_ \_ to \$ 38. 20 . \\
8 & Flight Centre \textbf{'s} \_ \_ \_ \_ \_ \_ to \$ 38. 20 . \\
9 & Flight Centre 's \textbf{shares} \_ \_ \_ \_ \_ to \$ 38. 20 . \\
10 & Flight Centre 's shares \textbf{rose} \_ \_ \_ \_ to \$ 38. 20 . \\
11 & Flight Centre 's shares rose \textbf{by} \_ \_ \_ to \$ 38. 20 . \\
12 & Flight Centre 's shares rose by \_ \_ \textbf{yesterday} to \$ 38. 20 . \\
13 & Flight Centre 's shares rose by \textbf{3} \_ yesterday to \$ 38. 20 . \\
14 & Flight Centre 's shares rose by 3 \textbf{cents} yesterday to \$ 38. 20 . \\
(target) & Flight Centre shares were up 3c at \$ 38.20 yesterday . \\

\bottomrule
\end{tabular}

\caption{Example sentences generated following an {\color{blue} Right-to-Left-to-Right} generation order, using the \textbf{learned} decoding algorithm on De$\rightarrow$En.}
    \label{fig:learned-gen-order-ex-1}
\end{figure*}

\begin{figure*}[t!]
    \centering
    \centering
\footnotesize
\begin{tabular}{cc}
%\begin{tabular}{cccccccccccccccccccc}
\toprule
Iteration & {\color{red} Outside-In} \\
\midrule
(source) & Terminal 3 wird vor allem von kleineren US-Fluggesellschaften bedient .\\
1 & \_ \_ \_ \_ \_ \_ \_ \_ \_ \textbf{.} \\
2 & \textbf{Terminal} \_ \_ \_ \_ \_ \_ \_ \_ . \\
3 & Terminal \textbf{3} \_ \_ \_ \_ \_ \_ \_ . \\
4 & Terminal 3 \_ \_ \_ \_ \_ \_ \textbf{airlines} . \\
5 & Terminal 3 \_ \_ \_ \_ \_ \textbf{US} airlines . \\
6 & Terminal 3 \_ \_ \_ \_ \textbf{smaller} US airlines . \\
7 & Terminal 3 \_ \_ \_ \textbf{by} smaller US airlines . \\
8 & Terminal 3 \textbf{is} \_ \_ by smaller US airlines . \\
9 & Terminal 3 is \textbf{mainly} \_ by smaller US airlines . \\
10 & Terminal 3 is mainly \textbf{served} by smaller US airlines . \\
(target) & Terminal 3 serves mainly small US airlines . \\

\bottomrule
\end{tabular}

\caption{Example sentences generated following an {\color{red} Outside-In} generation order, using the \textbf{learned} decoding algorithm on De$\rightarrow$En.}

    \label{fig:learned-gen-order-ex-2}
\end{figure*}

\begin{figure*}[t!]
    \centering
    \centering
\footnotesize
\begin{tabular}{cc}
%\begin{tabular}{cccccccccccccccccccc}
\toprule
Iteration & {\color{green} Left-to-Right} \\
\midrule
(source) & Die Gewinner des Team- und Einzelwettkampfs erhalten Preise .\\
1 & \_ \_ \_ \_ \_ \_ \_ \_ \_ \_ \_ \textbf{.} \\
2 & \textbf{The} \_ \_ \_ \_ \_ \_ \_ \_ \_ \_ . \\
3 & The \textbf{winners} \_ \_ \_ \_ \_ \_ \_ \_ \_ . \\
4 & The winners \textbf{of} \_ \_ \_ \_ \_ \_ \_ \_ . \\
5 & The winners of \textbf{the} \_ \_ \_ \_ \_ \_ \_ . \\
6 & The winners of the \textbf{team} \_ \_ \_ \_ \_ \_ . \\
7 & The winners of the team \textbf{and} \_ \_ \_ \_ \_ . \\
8 & The winners of the team and \textbf{individual} \_ \_ \_ \_ . \\
9 & The winners of the team and individual \textbf{competitions} \_ \_ \_ . \\
10 & The winners of the team and individual competitions \textbf{will} \_ \_ . \\
11 & The winners of the team and individual competitions will \_ \textbf{prizes} . \\
12 & The winners of the team and individual competitions will \textbf{receive} prizes . \\
(target) & The winners of the team and individual contests receive prizes . \\

\bottomrule
\end{tabular}

\caption{Example sentences generated following an {\color{green} left-to-right} generation order, using the \textbf{learned} decoding algorithm on De$\rightarrow$En.}
    \label{fig:learned-gen-order-ex-0}
\end{figure*}

\end{document}